# Distinguishing Transformative from Incremental Clinical Evidence: A Classifier of Clinical Research using Textual features from Abstracts and Citing Sentences

Xuanyu Shi, Jian Du

National Institute of Health Data Science at Peking University

Beijing, 100191, China

Corresponding author: Jian Du, dujian@bjmu.edu.cn

**Abstract**

In clinical research and clinical decision-making, it is important to know if a study changes or only supports the current standards of care for specific disease management. We define such a change as transformative and a support as incremental research. It usually requires a huge amount of domain expertise and time for humans to finish such tasks. Faculty Opinions provides us with a well-annotated corpus on whether a research challenges or only confirms established research. In this study, a machine learning approach is proposed to distinguishing transformative from incremental clinical evidence. The texts from both abstract and a 2-year window of citing sentences are collected for a training set of clinical studies recommended and labeled by Faculty Opinions experts. We achieve the best performance with an average AUC of 0.755 (0.705-0.875) using Random Forest as the classifier and citing sentences as the feature. The results showed that transformative research has typical language patterns in citing sentences unlike abstract sentences. We provide an efficient tool for identifying those clinical evidence challenging or only confirming established claims for clinicians and researchers.

**Keywords**

clinical study, knowledge discovery, natural language processing, citing sentences analysis, machine learning

## 1. Introduction

According to whether puzzle-solving or paradigm-breaking, scientific research can be divided into two types: incremental research in normal science and transformative research leading to scientific revolution (Kuhn, 1970). The former is to supplement and develop further the established research under the existing scientific paradigm; the latter refers to such research that shifts or disrupts the established scientific paradigm. Transformative research involves ideas, discoveries, or tools that challenge current understanding or milestone breakthroughs which provide pathways to new frontiers. Transformative research often contains controversial results and thus may ultimately contradict prevailing wisdom. Identifying potential transformative research early and accurately is important for funding agencies and policymakers to maximize the impact of their investment. It also helps scientists identify and focus their attention on uncertain, contradictory, or controversial scientific issues and thus improve their research merit. However, operationalizing transformative research is challenging. If an evaluation is sought too soon, the transformative nature of the results may not be evident enough to be observed.

Nevertheless, distinguishing transformative research from incremental research has attracted more and more interest in recent years. Currently, citation network analysis and Natural Language Processing (NLP) on citation sentences (or citances) are leveraged for detecting transformative research. One of the representative works is the "disruptiveness" measure for a given article in its three-generational citation network (Wu, Wang, & Evans, 2019). The inter-generational model of citation network consists of the focus article, its cited references (i.e., the parent-generation), and the citing articles (i.e., the children-generation). When future citations of the article do not cite its references, i.e., they do not acknowledge the article's own intellectual forebears, then the extent of disruptiveness of the focus article is higher. Since the work done on its basis does not refer to the work on which it is based, we can speculate that it may open up a new perspective of the concerned research problem. This approach takes into account external (structural) features of the inter-generational citation network. The internal feature is also used for detecting transformative research. Such work often focused on citation context, i.e., the text surrounding citations, which provide a rich source of information on how citing authors characterize and utilize earlier literature, and describe the contributions. In our previous study, we have proposed an idea on mining both authors' claims and citing sentences for identifying potential transformative knowledge (Du, Li, Haunschild, Sun, & Tang, 2020). It is possible to provide a list of transformative research discoveries from the perspectives of both author's claims (e.g., the authors assert that they disprove previously published data or hypotheses) and the science community's comments (e.g., the given research challenges established dogma).

How to distinguish transformative from incremental research? Here we put this problem in the context of clinical medicine, especially focusing on clinical trials. Clinical evidence is predominantly disseminated in unstructured natural language scientific manuscripts that describe the conduct and results of randomized control trials (RCTs) (Lehman, DeYoung, Barzilay, & Wallace, 2019). With the ever-growing number of clinical evidence, for example, seventy-five trials and eleven systematic reviews a day, how will we ever keep up (Bastian, Glasziou, & Chalmers, 2010)? One solution is classifying massive medical evidence: is it transformative or incremental, i.e., is it changing or only supporting current standards of care for a specific disease? This provides clinicians and decision-makers an enhancement of efficiency in the clinical decision support system. When we are reading clinical research related studies (in this paper we use 'study/studies', 'research/researches', and 'article/articles' as synonyms), it is informative and time-saving if we can differentiate whether a study is transformative or only incremental to existing scientific assertions.

Labeling studies as transformative or incremental requires strong medical knowledge background. Fortunately, Faculty Opinions provides recommended studies selected by top researchers in the biological and medical field. The website https://facultyopinions.com/, originally called "Faculty of 1000" (F1000) is a platform that provides a service of post-publication peer reviews by the top scientists and physicians (P. Wang, Williams, Zhang, & Wu, 2020). It gives people a high-level view with an understanding of the most recommended articles in the field. Experts recommend articles with a detailed explanation and corresponding category tags, such as "changes clinical practice", "refutation", "controversial", "confirmation" and so on. In other words, Faculty Opinions provides us with a well-annotated corpus about whether a research challenges or only confirms established research. The annotation criteria in this study are based on information collected from this website.

Natural Language Processing (NLP) is a useful technique in accessing and using health-related information to support clinical decisions (Demner-Fushman, Chapman, & McDonald, 2009). It helps health care providers and researchers to discover latent knowledge and medical relationships for facilitating clinical studies and solving clinical problems (Y. Wang, Tafti, Sohn, & Zhang, 2019). Among which, citing sentences are comments from the scientific community who cite the articles, and contain more objective commentary descriptions, compared to abstracts that are written by the authors themselves. Besides measuring the impact of studies, citation analysis helps researchers to discover what peers think of the studies.

Based on the functions that citing sentences played in evidence appraisal, our primary goal in this study is to replicate the experts' process of discovering transformative clinical studies. Reading articles and extracting information are time-consuming tasks for healthcare providers. Here we build an automatic classification system that distinguishes transformative from incremental clinical studies. We leverage recommendation category tags by Faculty Opinions experts as ground truth and use texts from citing sentences, abstracts, and a combination of both as feature options.

In this study, we will extract linguistic features using NLP tools to build machine learning classifiers to distinguish transformative from incremental clinical evidence. The proposed method can provide an efficient tool for identifying those clinical evidence challenging or only confirming established claims for clinicians and researchers. We are not planning to annotate each sentence manually, instead, we try to leverage a well-annotated corpus provided by Faculty Opinions, which were not fully utilized so far.

## 2. Related Work

Citing sentences analysis may help to identify incremental research as well as transformative research by classifying citation motivation as positive, negative, and neutral. Several attempts to classify the opinion an author holds toward a work that they cite (e.g., positive/negative attitudes or approval/disapproval) have been presented. Generally, if a citation is being used to highlight a gap or deficiency in the referenced work, then the language used will be suggestive of conflict relations between the two; if a citation is being used to back up the current work, then there are likely argumentative support relations between the two (Lawrence & Reed, 2020). It is useful to know which arguments were confirmed and accepted by the research community and which ones were disputed or even rejected. In a study (Radev & Abu-Jbara, 2012), the authors believe that analyzing citation text helps identify these controversial arguments automatically. In another study (Bertin, Atanassova, Sugimoto, & Lariviere, 2016), the authors used n-grams to study the location of negative and positive citations, and showed that that the word "disagree*" was much less likely to occur than the word "agree*", irrespective of papers' sections. Some researchers classified over 750,000 references made by papers published in the *Journal of Immunology* as either positive or negative, finding that negative citations tend to come in the first few years after the original study was published (Catalini, Lacetera, & Oettl, 2015). Negatively-cited studies are in the minority, but are usually high quality and receive more attention than the average. Leveraging a massive collection of scientific texts, some researchers (Lamers et al., 2021) developed a cue-phrase-based approach to identify instances of disagreement citations across more than four million scientific articles. A study was conducted to discover biomedical discoveries by applying citation context analysis (Small, Tseng, & Patek, 2017). They labeled citing sentences containing "discovery words" (discover*) as "discovery citance". Then they separated articles into three categories: "violation", "innovation" and "extension" by the research directions towards the earlier discoveries. Among these efforts, semantic lexical resources and NLP tools are used to create a network of opinion polarity relations. Sentences containing citations are extracted first, before determining the sentimental orientation through the subjective cue words in the context of the citation. From these sentimental orientations, the attitude of the author toward the work that they are citing is labeled.

Recently, citing sentences have been used to represent a linkage between a piece of evidence and a claim for argumentation mining (Wadden et al., 2020). They constructed an expert-annotated dataset of 1,409 scientific claims accompanied by abstracts that support or refute each claim and annotated with rationales justifying each SUPPORTS / REFUTES decision. They cleverly take citing sentences as the naturally occurring claims in the scientific literature, and the cited references as the pieces of evidence. Using citing sentences as a source of claims speeds up the claim generation process. In addition, citation links indicate the exact documents likely to contain necessary evidence to verify a given claim. Claims are natural since they are derived from citing sentences, which are formed by the science community.

There are few studies focusing on the citing sentences of clinical trials. In one study,

the authors used machine learning methods to classify citation sentiments of clinical trial papers (J. Xu et al., 2015). They annotated the citing sentences as *positive*, *neutral,* and *negative*. They extracted features including N-grams, sentiment lexicons, and structure features and they achieved a highest Micro-F score of 0.860 and a Macro-F score of 0.719. There was a serious problem of class imbalance in the study. They addressed this problem and proposed that semi-supervised learning could be a potential method to handle imbalanced data. Besides, they performed a comprehensive comparison between clinical studies and other fields in terms of citation context analysis, and concluded that it is challenging that in biomedical papers, researchers tend to use different linguistic patterns compared to other fields (J. Xu et al., 2015).

Based on Xu, et al.'s annotation corpus, some authors compared three different approaches for sentiment analysis of citing sentences in clinical trial publications (Kilicoglu et al., 2019). They concluded that a Convolutional Neural Network with hand-crated features yielded the best performance for sentiment classifications of citation texts (0.882 accuracy and 0.721 macro-$F_1$ score). Compared to the works by Kilicoglu, our goal is to classify if a clinical study is transformative or incremental to the prior clinical evidence, instead of classifying the sentimental orientation of its citation texts. Besides, we are going to use sentiment score as one type of feature and combine it with other linguistic features for classification.

F1000 recommends important publications in the biomedical literature, and from this perspective, F1000 could be an interesting tool for research evaluation of citation metrics (Waltman & Costas, 2014). In our previous study, we investigated the effects of research type on the differences between citation metrics and F1000 recommendations (Du, Tang, & Wu, 2016). We found that the nonprimary research (such as Review/Commentary) or evidence-based research articles (such as Systematic Review/Meta-Analysis, Clinical Trial: RCT, Clinical Trial: non-RCT, Confirmation, New Finding, and Technical Advance) were more likely than other types to be highly cited yet less highly recommended. By contrast, the translational or transformative research articles (e.g., Interesting Hypothesis, Controversial Topic, Novel Drug Target, Changes Clinical Practice, and Refutation) were more likely to be highly rated by peer reviewers yet less cited than articles with other labels. The results indicated that attempts to improve the application of bibliometric tools in research assessment procedures should consider the types of publications: (a) incremental (evidence-based) research publications; (b) transformative research or high-risk research publications (such as refutation, reversal or contradiction); and (c) translational research publications (such as indicating novel drug target, and changing clinical practice). Benefited from this well-annotated resource -- Faculty Opinions, here we turned to another question, i.e., can we distinguish transformative from incremental clinical evidence based on Faculty Opinions labels and machine learning techniques?

## 3. Data and Methods

We treat our research as a binary classification problem, distinguishing transformative (T) from incremental (I) clinical studies. Our corpus is built from the studies that are recommended by field experts on the https://facultyopinions.com/ website. We only include clinical studies (e.g., clinical trials and observational studies) by applying a search query on PubMed (a database of biomedical journal articles, developed by the National Center for Biotechnology Information (NCBI) at the National Library of Medicine (NLM)) (Canese & Weis, 2013) and non-clinical studies are excluded.

We propose a rule to group expert recommendation tags and then use the groups for labeling the studies. We also remove the studies with conflicting tags. For example, if a clinical study is labeled with both "changes clinical practice" and "confirmation" at the same time, it is hard to decide whether it refutes or supports earlier clinical evidence. Such ambiguous samples are excluded. Also, as an enhancement of annotations, we only include those studies with two or more expert recommendations.

We collect abstracts and citing sentences of the included studies and extract linguistic features. A feature matrix is constructed consisting of Part-of-speech tagging (POStags), sentiment scores, and N-gram transformations. With weights adjustment and ten-fold cross-validation, the features matrix is fed into binary machine learning classifiers, including Random Forest (RF) (Breiman, 2001), LightGBM (Ke et al., 2017), AdaBoost (Schapire, 2013), and XGBoost (Chen & Guestrin, 2016).

Figure 1 shows an overview of the workflow of the proposed approach.

## 3.1 Data Collection

### 3.1.1 Clinical Evidence

In this study, we use a well-annotated resource provided by Faculty Opinions. On the https://facultyopinions.com/ website, recommended studies have one or more corresponding category tags, labeled by domain experts. All the articles on the website have their corresponding recommendation tags by domain experts.

Faculty members tag articles with one or more of the following labels:

***Confirmation***: the article validates previously published data or hypotheses;

***Changes Clinical Practice***: the article recommends a complete, specific, and immediate change in practice by clinicians for a defined group of patients;

***Controversial***: the article challenges established dogma;

***Refutation***: the article disproves previously published data or hypotheses;

***Good for Teaching***: a key article in the field and/or one which is well written;

***Interesting Hypothesis***: the article presents a new model;

***New Finding***: the article presents original data, models, or hypotheses;

***Novel Drug Target***: the article suggests new targets for drug discovery; and

***Technical Advance***: the article introduces a new practical/theoretical technique, or a novel use of an existing technique.

In this study, we select the articles with the following specific recommendation tags as our training and testing corpus: Confirmation, Changes Clinical Practice, Refutation, and Controversial. Details are discussed in the annotation section.

Each study on the website has at least one recommendation by a field expert. It also has one or more recommendation tags that describe its characteristics or contribution. We acquire 44,813 studies in total with their PMIDs and corresponding recommendation tags by experts.

Next, the articles are filtered and limited to clinical studies, by using a search query using Medical Subject Headings (MeSH) provided by PubMed:

*((clinical[Title/Abstract] AND trial[Title/Abstract]) OR clinical trials[MeSH Terms] OR clinical trial[Publication Type] OR random\*[Title/Abstract] OR random allocation[MeSH Terms] OR therapeutic use[MeSH Subheading]).*

MeSH is a hierarchically-organized vocabulary thesaurus used for organizing medical information and indexing articles on PubMed (Lipscomb, 2000). After being limited to clinical studies, the size of the corpus decreases to 13,552. Abstracts of the articles are acquired using E-utilities (Entrez Programming Utilities) (Sayers, 2017) developed by NCBI (National Center for Biotechnology Information) (Bethesda (MD): National Library of Medicine (US), 1988).

### 3.1.2 Citing Sentences

Scientific commentaries and citing sentences are important when we appraise an article (Rogers, Mills, Grossman, Goldstein, & Weng, 2020). Citing sentences are acquired using Colil (Comments on Literature in Literature) (Fujiwara & Yamamoto, 2015). Colil is a literature database having citation contexts that are extracted from full-text papers of the PubMed Central Open Access Subset (PMC-OAS). In the Colil database, citation contexts are stored as RDF (Resource Description Framework) format. RDF is a language and a format for information representation about resources on the World Wide

Web (Manola, Miller, & McBride, 2004) and SPARQL is the standard query language for RDF format data (Pérez, Arenas, & Gutierrez, 2009). We use SPARQL queries to retrieve the sets of citation contexts of the articles from Colil.

## 3.2 Annotation of Corpus

### 3.2.1 Data labeling

We separate the four recommendation tags into two classes, transformative (T) and incremental(I), which will also be the labels for our data. Following are explanations of the classes:

*Transformative*: A study that is challenging established claims

*Incremental*: A study that is confirming established claims

We annotate the articles as *T* class if it is recommended with two or more tags from (*Refutation, Changes Clinical Practice, Controversial*), and as *I* class if it is recommended with the tag '*Confirmation*' from at least two experts. Figure 2 shows the corpus annotation based on recommendation tags by Faculty Opinions experts.

### 3.2.2 Conflicting labels removal

A study can be recommended with tags of both classes, for example, *Confirmation* and *Controversial* at the same time. Such a recommendation is contradictory and ambiguous. We remove such studies (n=318) to increase the quality and certainty of the annotation.

Since studies that are recommended by only one expert may be somewhat subjective, we enhance the annotations by only including the studies with two or more experts' recommendations tags from the same labeling class to minimize the bias as much as possible. Finally, 913 articles are left for further analysis. after the enhancement.

## 3.3 Features

The purpose of this study is to discover transformative clinical studies in an early stage. Thus, we filter out the citing sentences that are later than the first two years of publication. Then we combine all the remaining earlier 2-year citing sentences and joint them as a long text. Besides citing sentences, we acquire the text of the abstract for each study using E-utilities developed by NCBI. We will compare the performances with different textual combinations from citing sentences and abstracts.

There are three types of features extracted from the texts.

1. Part-of-speech tagging (POS tags)

Part-of-speech tagging is of process of transforming a sentence or a text into a sequence of revised and shortened lexical terms such as (Noun, NN), (Verb, VB), (Determiner, DT), etc. We extract POS tags features using a Python NLP package Spacy (Honnibal & Montani, 2017). POS tags represent the syntactic attributes of a word, such as verbs, nouns, adjectives, plurals, etc (Balwant, 2019). As the researchers discovered in their research on sentiment analysis using Twitter as a corpus (Pak & Paroubek, 2010), they observed that there exists specific patterns of POS tags in different kinds of natural language expressions. For example, 'verbs, non-3$^{rd}$ person singular present' (VBP) are more often used by authors of subjective opinions. And 'comparative adjectives' (JJR) are used for comparison and presenting facts. Also, 'verbs in past tense' (VBN, VBD) appear more in negative expressions.

2. Sentiment score

Sentiment scores are acquired by using the Vader (Valence Aware Dictionary for sEntiment Reasoning) sentiment analysis tool (Hutto & Gilbert, 2014) managed by NLTK (Loper & Bird, 2002) (version 3.6.2) and the compound scores of sentiment polarity scores are saved as our feature. Vader uses a rule-based model for sentiment score calculation by developing sentimental lexicons. In this study, we select the sentimental compound score, proposed by NLTK, as the sentiment score. The sentimental compound score is calculated by summing up the normalized lexicon scores. This rule-based model is not only computationally economical, but also inspectable, well-explained and accessible compared to complex machine learning classifiers like SVM (support vector machine) (Hutto & Gilbert, 2014). Sentiment scores give an instant and direct sense of whether a text's sentiment is positive or negative. Nevertheless, in the field of clinical studies, we need more sources and types of features to inspect in terms of classifying texts.

3. N-grams

In natural language, N-grams are sequences of words of length N extracted from sentences (Fürnkranz, 1998). Uni-grams, bi-grams, and tri-grams represent sequences of words of length 1, 2, and 3 respectively. We use the tool CountVectorizer and TfidfVectorizer developed by Scikit-learn (Pedregosa et al., 2011) (version 0.24.2) to automatically generate N-grams from the corpus and transform them into matrices. CountVectorizer generates a sparse representation of word counts while TfidfVectorizer generates TF-IDF scores. For explanation, TF-IDF (Term Frequency-Inverse Document Frequency) represents the relevancy of a specific term to its document. TF-IDF score of a term *t* in a document *D* is calculated by the formula (1):

$$\text{TF} - \text{IDF}(t, d) = \text{TF}_{t,d} \times \text{IDF}_{t,d,N} = tf_{t,d} \times \log\left(\frac{N}{df_t}\right) \qquad (1)$$

In the formula (1), $tf_{t,d}$ is the frequency of term *t* in document *d*. $df_t$ is the number of documents containing the term *t* in the whole corpus. And *N* is the total number of documents in the whole corpus. The TF-IDF technique is often used in the field of text mining because it can reflect the importance of a word to its corpus (Jing, Huang, & Shi, 2002). In this study, we extract N-grams from n=1 to n=3, in other words, all the uni-grams, bi-grams and tri-grams are extracted and saved for modeling. We will compare the performances between the usage of count score and TF-IDF score as a part of features. Table 1 demonstrates two example sentences and generated features.

**3.4 Machine Learning Classifiers**

Combining all the three types of features and transforming them into one matrix, we feed it into different binary machine learning classifiers and compare the performances.

1) *Weights Adjustment*: Since there exists an imbalance between classes of transformative vs incremental (the numbers of the two classes of dataset are unequal), we apply weights adjustment of the classes in the process of training classifiers in case of results leaning towards the imbalanced class.

2) *Vectorizers*: In this study, we apply two text vectorizers by Scikit-learn (Pedregosa et al., 2011) to transform text data into Count and TF-IDF(term frequency-inverse document frequency, a measure of word importance in a document) (Kim & Gil, 2019) matrices as features:
    - *Countvectorizer*: building a matrix of tokens by counts from a collection of texts
    - *Tfidfvectorizer*: building a matrix of tokens by TF-IDFs from a collection of texts

3) *Classifiers*: Four different binary machine learning classifiers are used for classification and compared for performance: Random Forest, LightGBM, AdaBoost, and XGBoost. We select these tree-based algorithms such as Random Forest because of their simplicity and promising performance in solving high dimensional data problems such as text classification (B. Xu, Guo, Ye, & Cheng, 2012).

4) *Cross-Validation*: Due to the relatively small size of data and potential unequal distribution of data, we perform ten-fold cross-validation in the process of training and testing.

5) *Evaluation*: To evaluate our classification models, we choose to use Area Under the Receiver Operating Characteristic Curve (ROC AUC Score/AUC). ROC curve plots the True Positive Rate (TPR) against the False Positive Ratio (FPR) of a classifier. The AUC number represents the ability of a classifier to rank positives ahead of negatives (Forman & Scholz, 2010). The higher the AUC score is achieved, the better the capability of the classifier is to distinguish between classes. Compared to the accuracy score, which is calculated by the number of correct classifications divided by the total number of classifications, the AUC score reflects a more reliable evaluation of the classifier. It focuses on the model itself rather than the data, predicted by a decision threshold. In other words, the AUC score gives an indication of how much work a classifier really does, with lower scores to the random selection or "one class only" classifiers (Bradley, 1997).

In cross-validations, (Forman & Scholz, 2010) compared two methods of handling metrics in AUC scores of the folds:

- *$AUC_{merge}$*: sorting scores from all folds into a single ROC curve then computing AUC
- *$AUC_{avg}$*: computing AUC scores of each fold and taking the average of the sum

In this study, we choose $AUC_{avg}$ because we intended to focus on comparing the abilities of the classifiers to distinguish between positives and negatives, also for the reason that $AUC_{merge}$ requires precise calibrations and specific threshold values for a reliable result (Forman & Scholz, 2010).

## 4. Results

### 4.1 Exploratory Data Analysis

During the acquisition of feature data, nine articles are removed because we are not able to acquire the abstracts from PubMed, and 161 articles are removed because they had not been cited in the first two years after publication (the reasoning is explained in the features section). Eventually, a corpus of 743 articles is built with N(incremental)=365 and N(transformative)=378. We conduct an exploratory data analysis. Table 2 compares the major MeSH terms of selected clinical studies, and Table 3 shows descriptive statistics of the studies.

**4.2 Classification Results**

The results of comparing different machine learning classifiers on multiple combinations of vectorizers/features are shown in Table 4. We achieve the highest $AUC_{avg}$ of 0.755 (0.705-0.875) using a random forest classifier with CountVectorizer (1 to 3)-grams transformations and citing sentences as feature.

Citing sentences model outperforms abstract model and citing sentences + abstract model. For comparison among the applied features, the highest achieved scores of citing sentences, abstract, and citing sentences+abstract are 0.755, 0.576, and 0.701, respectively. And for comparison between CountVectorizer and TfidfVectorizer, the highest $AUC_{avg}$ scores are 0.755 and 0.616, respectively.

From the model in which the highest $AUC_{avg}$ is achieved, we record the top 20 important features with importance scores from the machine learning model. The scores are calculated as the mean and the standard deviation of accumulation of the impurity decrease with each tree in tree algorithms using Scikit-learn (Pedregosa et al., 2011). The scores represent the impact of the features on the classification. From Table 5 we can see that most of the important features are POS tags while there are two uni-gram words, "however" and "question", and the sentimental compound score is also one of the top important features.

**4.3 Error Analysis**

After training and testing the machine learning classifiers, we conduct an error analysis to exhibit the difference in citing sentences and major MeSH terms of falsely classified studies. In this study, a false classification means false positive or false negative in the evaluation process of a classification model. We save the falsely classified studies in each iteration of cross-fold validations and combine them for comparison. In error analysis, first, we analyze the major MeSH terms to discover in which areas the transformative/incremental clinical trial studies are falsely classified.

Table 6 shows the comparison in major MeSH terms between correctly classified studies and falsely classified studies. In areas such as Critical Illness/therapy* and Respiration, Artificial*, studies tend to be classified falsely in terms of whether they are transformative or incremental. In contrast, in areas such as Anti-Bacterial Agents/therapeutic use* and Antineoplastic Combined Chemotherapy Protocols/therapeutic use*, such studies are more easily classified correctly.

We also manually observe some examples of citing sentences of falsely classified articles to discover potential reasons. Table 7 shows each two of the correctly and falsely classified studies with their corresponding classes, PubMed IDs, and citing sentences. We find that it is a hard task to distinguish between the classes by only observing the citing sentences with manual judgment. While our proposed machine learning model is able to classify the classes with a decent $\text{AUC}_{\text{avg}}(0.755)$, we believe the model is meaningful that it can discover latent patterns and use them for accurate classifications.

## 5. Discussion and Conclusion

### 5.1 Summary

We have connected human knowledge in Faculty Opinions with machine learning methods to solve the problem of identifying transformative clinical studies. After all, in the field of clinical researches, highly professional knowledge is required to complete such a task. In this study, multiple binary machine learning classifiers were built to distinguish transformative from incremental clinical studies. We achieved an $\text{AUC}_{\text{avg}}$ of 0.755 (0.705-0.875) with a CountVectorizer and citing sentences as the feature. The interval between the lowest AUC score and the highest AUC score is 0.705-0.875 from the ten-fold cross validations. CountVectorizer outperforms TfidfVectorizer overall from the results. The reason behind this may be that TF-IDF scores reflect keywords of articles while word frequencies discover overall word and linguistic patterns. In this study, keywords (TF-IDF) do not play a decisive role in the classifications while word patterns (Count) help the classifiers achieve higher scores.

We also discover that linguistic patterns in citing sentences are more important and supportive than in abstracts when distinguishing transformative from incremental clinical studies. In other words, abstract text provided by authors themselves seems a noisy signal. We think citing sentences represent the comments on a given study from the scientific community, while the text in the abstract only indicates the author's individual statement to his/her research. In this study, a successful classifier based on citing sentences of the clinical trial studies help us conclude that opinions from the scientific community play an important role in discovering transformative clinical studies, and we believe in the huge potential of citing sentences in clinical evidence discovery. Compared to the research done by (J. Xu et al., 2015) in sentiment analysis of citing sentences of clinical trial studies, we accomplish a task of distinguishing transformative from incremental clinical studies using POS tags, N-grams, and sentiment scores as features.

It is also worth mentioning again that we only use the first two years of citing sentences after the publication date. With this model, we can discover transformative clinical studies at an early stage and provide the discovered studies to health care providers for applications such as selecting the state-of-the-art clinical evidence for clinical decision support, knowledge graph construction, and patient education. (Small et al., 2017) also mentioned the timing of recognition in terms of counting the interval between the publication year and the year when the earliest "discovery citance" occurs. A notable example (Small et al., 2017) is that an xenotropic murine leukemia virus-related virus (XMRV) study accumulated 26 citations in its first four years after publication in 2006, with 5 of 14 citations (36%) labeled as discovery in 2009, followed by 13 of 109 (12%) in 2010. The number started to decrease rapidly afterward because of questions about contamination. And after its retraction for the above reason in 2012, there were still 4 of 32 (12%) citations calling the study a discovery, showing a lag in recognition of medical studies. In this study, instead of using all citing sentences for a given paper, we use only a 2-year window of citing sentences as early as possible after its publication. We try to capture the earliest commentary textual feature (if any) for classifying clinical studies as transformative or incremental.

The results showed that transformative research has typical language patterns in citing sentences. It is the community of citing authors that ultimately decides what is and is not a transformative research, and this community designation extends over a number of years and is potentially subject to revision, as in the case of retractions. However, if a transformative research is overlooked or goes unrecognized by the community, it will be missed by citing sentences analysis. We can use citing sentences to discover transformative research (different classifiers giving highest AUC scores near 90%).

Citing sentences may not be so sensitive to scientific evaluation on a given work compared with the sentences from published commentaries, which are communications that facilitate reactions in response to published literature. Examples include letters to the editor and editorials. Most recently, (Rogers et al., 2020) provided an original overview of published commentary on clinical research articles in PubMed. They found a low prevalence of commentary as a mechanism for evaluating published studies, as less than 5% of clinical research studies had at least one comment. In general, the sentiment of clinical research commentaries demonstrates an overall supportive tone. While once dividing comments into editorials and letters, they implicated a different sentimental orientation. For example, editorials, which were often published at the same time as the commented original article, generally presented the article in the context of other evidence in a relatively supportive manner. In contrast, letters displayed evidence of communicating concerns about the published study, such as the lack of clarity in variable definitions or doubts regarding statistical assumptions. When a letter is potentially published, it can be accompanied by responses from the study authors, which usually involve addressing concerns brought up by the letter. So, any analysis on scientific commentary needs to take the comment type into careful consideration.

To train a classifier model, we extracted and used as textual features all types of words/phrases, including content words/phrases (e.g., medical/technical terms) and cue phrases (e.g., adjectives, adverb, action verb, conjunction). We want to include as much information as we can for both content-wise and sentiment-wise analysis. We intend to include not only topic independent features like cue phrases, but also subject-related features into our model to discover the frequency differences among the different subjects/areas/topics in which the transformative studies are produced. This is more important in clinical studies and can help researchers with which research areas they should pay more attention to. The combination of both content and cue words gives us a decent result, but cue phrases play a more important role in this classification task. As is shown in Table 5 we can see that most of the important features are verb, adjective, determiner, adverb tags as well as such words as "however" and "question", which are all topic independent cue words/phases. And in the future, we will explore the performance of the classifier by masking the content phrases such as the diseases, drugs, genes related entities to emphasize only the cue words/phases that are topic independent.

**5.2 Limitations**

Nevertheless, the number of strong and credible annotations is limited in our corpus. The goal of this study is to propose an automatic method to replicate the experts' process of selecting transformative clinical research. For continuously improving the performance of the classifier, we still need much more studies that are recommended by experts.

Another limitation is that we only acquired a limited number of citing sentences since the PubMed Central database only includes open access publications. As of 2017, PubMed Central Open Access Subset (PMCOAS) only covered 5.31% of all publications in the PubMed database (He & Li, 2019). Due to the incomprehensiveness of citing sentences, we did not use all the citing sentences of the studies as feature in our machine learning classifiers.

Besides, because the Faculty Opinions website only contains relatively high-level articles, we could not avoid the possibility that normal-level articles that are not recommended by the experts are also actually transformative.

**5.3 Future work**

This study is not only to use machine learning algorithms to dig textual features for transformative or incremental studies classification but also to explore the reasoning behind medical knowledge discovery and management. With the results we achieve in this study, we will be able to extend our research to using these transformative studies to help recognize state-of-the-art medical concepts and solutions and we can construct the most up-to-date clinical knowledge graphs based on these findings in the future.

**Acknowledgment**

This work was supported by the National Natural Science Foundation of China [72074006]; Peking University Health Science Center [BMU2021YJ008]; PKU-Baidu Fund [2020BD032]; and the Young Elite Scientists Sponsorship Program by China Association for Science and Technology [2017QNRC001].

# Figures and Tables

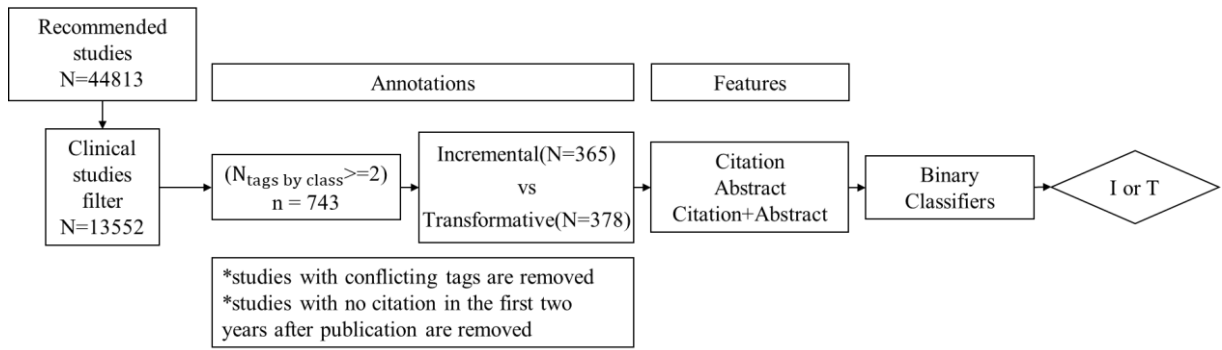

Figure 1 The workflow of the proposed classification method

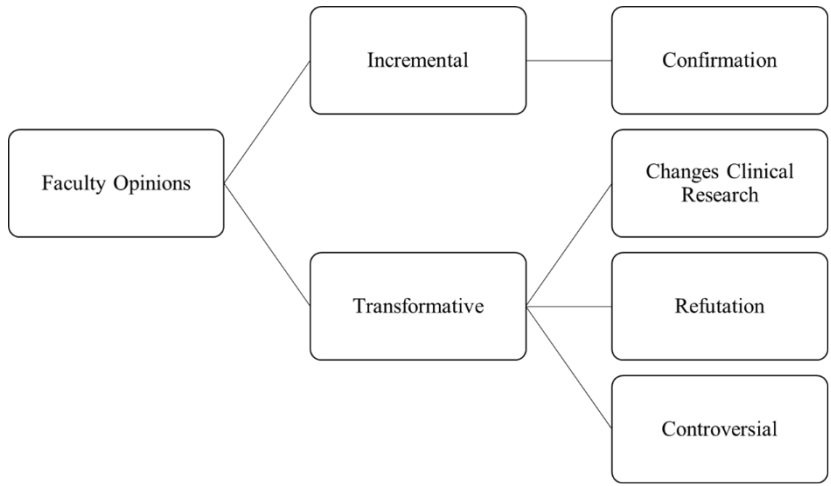

Figure 2 Annotation of classes by recommendation tag

**Table 1** Sample sentences and generated features

| Example sentence #1 | Feature name | Features |
|---|---|---|
| **This is supported by the recent AspECT clinical trial which found aspirin intake may have chemopreventative effects in Barrett's patients.** | POS tags | ('this', 'DT'), ('be', 'VBZ'), ('support', 'VBN'), ('by', 'IN'), ('the', 'DT'), ('recent', 'JJ'), ('AspECT', 'NNP'), ('clinical', 'JJ') ... |
| | Sentiment score | neg': 0.0, 'neu': 0.892, 'pos': 0.108, 'compound': 0.3182 |
| | Uni-gram | 'This', 'is', 'supported', 'by', 'the', 'recent', 'AspECT', 'clinical', 'trial'... |
| | Bi-gram | 'This is', 'is supported', 'supported by', 'by the', 'the recent', 'recent AspECT', 'AspECT clinical', 'clinical trial'... |
| | Tri-gram | This is supported', 'is supported by', 'supported by the', 'by the recent', 'the recent AspECT', 'recent AspECT clinical', 'AspECT clinical trial', 'clinical trial which'... |
| **Example sentence #2** | **Feature name** | **Features** |
| **However, despite these data, evidence to suggest failure in secondary prevention of CRC by total colonoscopy and polypectomy is emerging.** | POS tags | (however, 'RB'), (despite, 'IN'), (these, 'DT'), (data, 'NNS'), (evidence, 'NN'), (to, 'TO'), (suggest, 'VB'), (failure, 'NN'), (in, 'IN'), (secondary, 'JJ') ... |
| | Sentiment score | 'neg': 0.148, 'neu': 0.852, 'pos': 0.0, 'compound': -0.5106 |
| | Uni-gram | 'however', 'despite', 'these', 'data', 'evidence', 'to', 'suggest', 'failure', 'in', 'secondary'... |
| | Bi-gram | 'however despite', 'despite these', 'these data', 'data evidence', 'evidence to', 'to suggest', 'suggest failure' … |
| | Tri-gram | 'however despite these', 'despite these data', 'these data evidence', 'data evidence to' … |

**Table 2** Major MeSH terms comparison between Incremental and Transformative clinical studies

| Incremental (n=365) | | Transformative (n=378) | |
|---|---|---|---|
| major MeSH terms | count | major MeSH terms | count |
| Anti-Bacterial Agents/therapeutic use* | 9 | Anti-Bacterial Agents/therapeutic use* | 10 |
| Antineoplastic Combined Chemotherapy Protocols/therapeutic use* | 8 | Immunosuppressive Agents/therapeutic use* | 6 |
| Antibodies, Monoclonal/therapeutic use* | 6 | Hydroxymethylglutaryl-CoA Reductase Inhibitors/therapeutic use* | 5 |
| Propofol/administration & dosage* | 5 | Cross Infection/prevention & control* | 5 |
| Critical Illness/therapy* | 5 | Hospital Mortality* | 5 |
| Angioplasty, Balloon, Coronary* | 4 | Antineoplastic Combined Chemotherapy Protocols/therapeutic use* | 5 |
| Anti-Inflammatory Agents, Non-Steroidal/adverse effects* | 4 | Gastrointestinal Agents/therapeutic use* | 5 |
| Critical Care/methods* | 4 | Fluid Therapy/methods* | 5 |
| Cardiac Surgical Procedures* | 4 | Probiotics/therapeutic use* | 4 |
| Anti-Arrhythmia Agents/therapeutic use* | 4 | Cardiac Surgical Procedures* | 4 |

**Table 3** Descriptive statistics comparison between Incremental and Transformative clinical studies

|  | Incremental (n=365) | | Transformative (n=378) | |
|---|---|---|---|---|
|  | count | percentage | count | percentage |
| **Year** | | | | |
| *2001-2004* | 4 | 0.011 | 4 | 0.011 |
| *2005-2008* | 94 | 0.258 | 110 | 0.291 |
| *2009-2012* | 158 | 0.433 | 125 | 0.331 |
| *2013-2016* | 84 | 0.230 | 103 | 0.272 |
| *2017-2019* | 25 | 0.068 | 36 | 0.095 |
| **Common journals** | | | | |
| *The New England Journal of Medicine* | 43 | 0.118 | 69 | 0.183 |
| *The Lancet* | 26 | 0.071 | 21 | 0.056 |
| *The Journal of the American Medical Association* | 19 | 0.052 | 24 | 0.063 |
| *Anesthesiology* | 11 | 0.030 | 10 | 0.026 |
| *American Journal of Respiratory and Critical Care Medicine* | 11 | 0.030 | 9 | 0.024 |
| *Critical Care Medicine* | 9 | 0.025 | 5 | 0.013 |
| *Circulation* | 9 | 0.025 | 5 | 0.013 |
| *BMJ* | 8 | 0.022 | 8 | 0.021 |
| *Anesthesia and Analgesia* | 8 | 0.022 | 8 | 0.021 |
| **Number of comments** | | | | |
| *mean* | 2.6 | | 4.2 | |
| *median* | 2.0 | | 3.0 | |
| **Number of citances** | | | | |
| *mean* | 19.8 | | 42.9 | |
| *median* | 9.0 | | 14.0 | |

Table 4 Machine learning performance comparison

| CountVectorizer | Citing sentences | Abstract | Citing sentences+abstract |
|---|---|---|---|
| **RF** | **<u>0.755 (0.705-0.875)</u>** | 0.575 (0.426-0.663) | 0.701 (0.620-0.790) |
| **LGBM** | 0.716 (0.620-0.833) | 0.545 (0.472-0.620) | 0.691 (0.620-0.780) |
| **Ada** | 0.669 (0.584-0.826) | 0.536 (0.452-0.623) | 0.658 (0.560-0.724) |
| **XGB** | 0.716 (0.663-0.791) | 0.527 (0.419-0.620) | 0.673 (0.611-0.740) |
| **TfidfVectorizer** | **Citing sentences** | **Abstract** | **Citing sentences+abstract** |
| **RF** | 0.607 (0.551-0.676) | 0.576 (0.506-0.665) | 0.616 (0.562-0.663) |
| **LGBM** | 0.609 (0.529-0.665) | 0.539 (0.471-0.606) | 0.604 (0.512-0.692) |
| **Ada** | 0.583 (0.474-0.698) | 0.576 (0.508-0.622) | 0.613 (0.510-0.672) |
| **XGB** | 0.615 (0.500-0.663) | 0.535 (0.431-0.632) | 0.609 (0.546-0.709) |

Table 5 Features importance ranking

| Feature | Explanation | Importance Score |
| --- | --- | --- |
| **VBP** | verb, non-3rd person singular present | 1.081% |
| **MD** | verb, modal auxiliary | 0.863% |
| **JJ** | adjective (English) | 0.781% |
| **VBD** | verb, past tense | 0.775% |
| **DT** | determiner | 0.728% |
| **NNS** | noun, plural | 0.711% |
| **VBN** | verb, past participle | 0.657% |
| **RB** | adverb | 0.624% |
| **NNP** | noun, proper singular | 0.604% |
| **VBZ** | verb, 3rd person singular present | 0.588% |
| **IN** | conjunction, subordinating or preposition | 0.576% |
| **NN** | noun, singular or mass | 0.572% |
| **CC** | coordinating conjunction | 0.551% |
| **VB** | verb, base form | 0.529% |
| **however** | | 0.506% |
| **JJR** | adjective, comparative | 0.487% |
| **question** | | 0.483% |
| **VBG** | verb, gerund or present participle | 0.460% |
| **compound_score** | | 0.451% |
| **PRP** | pronoun, personal | 0.432% |

Table 6 Major MeSH terms comparison between correctly and falsely classified studies

| Correctly classified (n=556) | | Falsely classified (n=187) | |
| --- | --- | --- | --- |
| major MeSH terms | count | major MeSH terms | count |
| Anti-Bacterial Agents/therapeutic use* | 9 | Critical Illness/therapy* | 4 |
| Antineoplastic Combined Chemotherapy Protocols/therapeutic use* | 8 | Respiration, Artificial* | 3 |
| Antibodies, Monoclonal/therapeutic use* | 4 | Anticoagulants/administration & dosage* | 2 |
| Angioplasty, Balloon, Coronary* | 4 | Anemia/drug therapy* | 2 |
| Anti-Arrhythmia Agents/therapeutic use* | 4 | Atrial Fibrillation/drug therapy* | 2 |
| Endarterectomy, Carotid* | 4 | Antibodies, Monoclonal/therapeutic use* | 2 |
| Propofol/administration & dosage* | 4 | Blood Preservation* | 2 |
| Antihypertensive Agents/therapeutic use* | 4 | Placebo Effect* | 2 |
| Pemphigus/drug therapy* | 3 | Antifungal Agents/therapeutic use* | 2 |
| Anti-Inflammatory Agents, Non-Steroidal/adverse effects* | 3 | Critical Care/methods* | 2 |

Table 7 Citing sentences examples of falsely classified studies

| Truth | Classification | PMID | Citing sentence example |
|---|---|---|---|
| 1 | 0 | 16765760 | according to grgić et al. and crowther et al. , the complications related to neonatal respiratory system have noticeably decreased when steroids were used to the neonates born prior to 32+0-33+6 weeks. |
| 1 | 0 | 24268395 | with a side-effect and quality-of-life profile comparable to rfa13, 1-year anatomical success rates of between 70 per cent22 and 75 per cent after foam sclerotherapy are lower than those obtained with endothermal technology. |
| 0 | 1 | 19524353 | of trials using these products suggests potential for each of them to have a therapeutic role,34 and a recent multicenter rct has demonstrated statistically significant symptomatic improvement in patients receiving bee pollen extract. |
| 0 | 1 | 17000910 | statin use has been reported to protect against postoperative af in observational studies as well as in a randomized, placebo-controlled trial, 26 and we also observed a trend toward a protective effect of preoperative statin use. |